\documentclass[11pt]{article}
\usepackage{coling2020}
\usepackage{times}
\usepackage{url}
\usepackage{latexsym}
\usepackage{caption,tabularx,booktabs,array}
\usepackage{pgfplotstable}
\usetikzlibrary{calc}
\usepackage[group-minimum-digits=4,detect-weight=true,detect-inline-weight=math,round-mode=places]{siunitx}
\usepackage{todonotes}
\usepackage{fontawesome}
\usepackage[utf8]{inputenc}

\usepackage{arabtex}

\usepackage{utf8}
\setcode{utf8}

\usepackage[utf8]{inputenc}
\usepackage[LGR,OT1]{fontenc}

\newcommand{\textgreek}[1]{\begingroup\fontencoding{LGR}\selectfont#1\endgroup}

\colingfinalcopy 

\title{{S}em{E}val-2020 Task 12: Multilingual Offensive Language Identification\\ in Social Media ({O}ffens{E}val 2020)}

\author{Marcos Zampieri\textsuperscript{1}, Preslav Nakov\textsuperscript{2}, Sara Rosenthal\textsuperscript{3}, Pepa Atanasova\textsuperscript{4}, Georgi Karadzhov\textsuperscript{5} \\ \vspace{2mm}
{\bf Hamdy Mubarak\textsuperscript{2}, Leon Derczynski\textsuperscript{6}, Zeses Pitenis\textsuperscript{7}, Çağrı Çöltekin\textsuperscript{8}} \\ 
\textsuperscript{1}Rochester Institute of Technology, USA, \textsuperscript{2}Qatar Computing Research Institute, Qatar\\
\textsuperscript{3}IBM Research, USA, \textsuperscript{4}University of Copenhagen, Denmark, \textsuperscript{5}University of Cambridge, UK \\
\textsuperscript{6}IT University Copenhagen, Denmark, \textsuperscript{7}University of Wolverhampton, UK \\ 
\textsuperscript{8}University of Tübingen, Germany
  \\
  {\tt marcos.zampieri@rit.edu} \\}

\date{}

\begin{document}

\maketitle

\begin{abstract}
    We present the results and the main findings of SemEval-2020 Task 12 on Multilingual Offensive Language Identification in Social Media (OffensEval-2020). The task included three subtasks corresponding to the hierarchical taxonomy of the OLID schema from OffensEval-2019, and it was offered in five languages: Arabic, Danish, English, Greek, and Turkish. OffensEval-2020 was one of the most popular tasks at SemEval-2020, attracting a large number of participants across all subtasks and languages: a total of 528 teams signed up to participate in the task, 145 teams submitted official runs on the test data, and 70 teams submitted system description papers.
\end{list} 
\end{abstract}

\section{Introduction}

\blfootnote{This work is licensed under a Creative Commons Attribution 4.0 International License. License details: \url{http://creativecommons.org/licenses/by/4.0/}.}

Offensive language is ubiquitous in social media platforms such as Facebook, Twitter, and Reddit, and it comes in many forms.
Given the multitude of terms and definitions related to offensive language used in the literature, several recent studies have investigated the common aspects of different abusive language detection tasks \cite{waseem2017understanding,wiegand2018overview}.
One such example is \textit{SemEval-2019 Task 6: OffensEval}\footnote{\url{http://sites.google.com/site/offensevalsharedtask/offenseval2019}}\cite{offenseval}, which is the precursor to the present shared task. OffensEval-2019 used the Offensive Language Identification Dataset (OLID), which contains over 14,000 English tweets annotated using a hierarchical three-level annotation schema that takes both the target and the type of offensive content into account \cite{OLID}. The assumption behind this annotation schema is that the target of offensive messages is an important variable that allows us to discriminate between, e.g.,~hate speech, which often consists of insults targeted toward a {\em group}, and cyberbullying, which typically targets {\em individuals}. A number of recently organized related shared tasks followed similar hierarchical models. Examples include HASOC-2019 \cite{hasoc2019} for English, German, and Hindi, HatEval-2019 \cite{hateval2019} for English and Spanish, GermEval-2019 for German \cite{germeval-2019}, and TRAC-2020 \cite{trac2020} for English, Bengali, and Hindi.

OffensEval-2019 attracted nearly 800 team registrations and received 115 official submissions, which demonstrates the interest of the research community in this topic. Therefore, we organized a follow-up, OffensEval-2020\footnote{\url{http://sites.google.com/site/offensevalsharedtask/home}} (SemEval-2020 Task 12), which is described in this report, building on the success of OffensEval-2019 with several improvements. In particular, we used the same three-level taxonomy to annotate new datasets in five languages, where each level in this taxonomy corresponds to a subtask in the competition: 

\begin{itemize}
\item Subtask A: Offensive language identification;
\item Subtask B: Automatic categorization of offense types;
\item Subtask C: Offense target identification.
\end{itemize}

\noindent The contributions of OffensEval-2020 can be summarized as follows:

\begin{itemize}
\item We provided the participants with a new, large-scale semi-supervised training dataset containing over nine million English tweets \cite{SOLID}.
\item We introduced multilingual datasets, and we expanded the task to four new languages: Arabic \cite{mubarak2020arabic}, Danish \cite{sigurbergsson2020offensive}, Greek \cite{pitenis2020}, and Turkish \cite{coltekin2020}. This opens the possibility for cross-lingual training and analysis, which several participants indeed explored.
\item Compared to OffensEval-2019, we used larger test datasets for all subtasks.
\end{itemize}

Overall, OffensEval-2020 was a very successful task. The huge interest demonstrated last year continued this year, with 528 teams signing up to participate in the task, and 145 of them submitting official runs on the test dataset. Furthermore, OffensEval-2020 received 70 system description papers, which is an all-time record for a SemEval task. 

The remainder of this paper is organized as follows: Section~\ref{sec:schema} describes the annotation schema. Section~\ref{sec:data} presents the five datasets that we used in the competition. Sections~\ref{sec:taskparticipation}-\ref{sect:turkish} present the results and discuss the approaches taken by the participating systems for each of the five languages. Finally, Section~\ref{sec:conclusion} concludes and suggests some possible directions for future work.

\section{Annotation Schema}
\label{sec:schema}
OLID's annotation schema proposes a hierarchical modeling of offensive language. It classifies each example using the following three-level hierarchy:

\vspace{2mm}

\noindent \textbf{Level A - Offensive Language Detection} \\
Is the text offensive (OFF) or not offesive (NOT)?

\begin{description}
\item \textbf{NOT}: text that is neither offensive, nor profane;
\item \textbf{OFF}: text containing inappropriate language, insults, or threats.
\end{description}
\vspace{2mm}
\noindent \textbf{Level B - Categorization of Offensive Language} \\
Is the offensive text targeted (TIN) or untargeted (UNT)?

\begin{description}
\item \textbf{TIN}: targeted insults or threats towards a group or an individual;
\item \textbf{UNT}: untargeted profanity or swearing. 
\end{description}

\vspace{2mm}
\noindent \textbf{Level C - Offensive Language Target Identification}\\ Who or what is the target of the offensive content?

\begin{description}
\item \textbf{IND}: the target is an individual, which can be explicitly mentioned or it can be implicit;
\item \textbf{GRP}: the target is a group of people based on ethnicity, gender, sexual orientation, religious belief, or other common characteristic; 
\item \textbf{OTH}: the target does not fall into any of the previous categories, e.g.,~organizations, events, and issues. 
\end{description}

\section{Data}
\label{sec:data}

In this section, we describe the datasets for all five languages: Arabic, Danish, English, Greek, and Turkish. All of the languages follow the OLID annotation schema and all datasets were pre-processed in the same way, e.g.,~all user mentions were substituted by \texttt{@USER} for anonymization. The introduction of new languages using a standardized schema with the purpose of detecting offensive and targeted speech should improve dataset consistency. This strategy is in line with current best practices in abusive language data collection~\cite{vidgen2020directions}. All languages contain data for subtask A, and only English contains data for subtasks B and C. The distribution of the data across categories for all languages for subtask A is shown in Table~\ref{tab:labels}, while Tables~\ref{tab:labelsB} and \ref{tab:labelsC} present statistics about the data for the English subtasks B and C, respectively. Labeled examples from the different datasets are shown in Table~\ref{T:examples}.

\begin{table}[!ht]
\centering
\scalebox{0.88}{
 \begin{tabular}{l*{6}{S[table-format=6.0]}}
\toprule
    & \multicolumn{3}{c}{\bf Training} &\multicolumn{3}{c}{\bf Test}\\
    \cmidrule(lr){2-4}\cmidrule(lr){5-7}
    \bf Language & {OFF} & {NOT} & {Total} 
                 & {OFF} & {NOT} & {Total} \\ \midrule
  English & 1448861 & 7640279 & 9089140 & 1090 & 2807 & 3897 \\
  Arabic   & 1589 & 6411 & 8000 & 402 & 1598 & 2000\\
  Danish  & 384 & 2577 & 2961 & 41 & 288 & 329 \\
  Greek   & 2486 & 6257 & 8743 & 425 & 1119 & 1544 \\
  Turkish & 6131 & 25625 & 31756 & 716 & 2812 & 3528 \\
\bottomrule 
\end{tabular}
}
\vspace{-2mm}
\caption{Subtask A (all languages): statistics about the data.}
\label{tab:labels}
\end{table}

\begin{table}[!ht]
\centering
\scalebox{0.88}{
 \begin{tabular}{l*{6}{S[table-format=6.0]}}
\toprule
    & \multicolumn{3}{c}{\bf Training} &\multicolumn{3}{c}{\bf Test}\\
    \cmidrule(lr){2-4}\cmidrule(lr){5-7}
    \bf Language & {TIN} & {UNT} & {Total} 
                 & {TIN} & {UNT} & {Total} \\ \midrule
  English & 149550 & 39424 & 188974 & 850 & 1072  & 1922 \\
\bottomrule 
\end{tabular}
}
\vspace{-2mm}
\caption{Subtask B (English): statistics about the data.}
\label{tab:labelsB}
\end{table}

\begin{table}[!ht]
\centering
\scalebox{0.88}{
 \begin{tabular}{l*{8}{S[table-format=6.0]}}
\toprule
    & \multicolumn{4}{c}{\bf Training} &\multicolumn{4}{c}{\bf Test}\\
    \cmidrule(lr){2-5}\cmidrule(lr){6-9}
    \bf Language & {IND} & {GRP} & {OTH} & {Total} 
                 & {IND} & {GRP} & {OTH} & {Total} \\ \midrule
  English & 120330 & 22176 & 7043 & 149549 & 580 & 190 & 80 & 850 \\
\bottomrule 
\end{tabular}
}
\vspace{-2mm}
\caption{Subtask C (English): statistics about the data.}
\label{tab:labelsC}
\end{table}

\begin{table*}[!ht]
\centering
\scalebox{0.88}{
\begin{tabular}{cp{11cm}ccc}
\toprule
 \bf Language & \bf  Tweet & \bf  A & \bf  B & \bf C \\
\midrule
English & This account owner asks for people to think rationally. & NOT & --- &  --- \\
Arabic  & \<لعنة الله عليك يا سباك يا جبان يابن الكلب.> & OFF & --- &  --- \\
        & \textit{Translation: May God curse you, O coward, O son of a dog.} & & \\
Danish  & Du glemmer {\O}steuropaer som er de v{\ae}rste & OFF & --- &  --- \\
        & \textit{Translation: You forget Eastern Europeans, who are the worst} & & \\
Greek   & \textgreek{Παραδέξου το, είσαι αγάμητη εδώ και καιρό...} & OFF & --- &  --- \\
        & \textit{Translation: Admit it, you’ve been unfucked for a while now...} & & \\
Turkish & Böyle devam et seni gerizekalı\newline{}\textit{Translation: Go on like this, you idiot} & OFF & --- &  --- \\
English & this job got me all the way fucked up real shit & OFF & UNT  & --- \\
English & wtf ari her ass tooo big & OFF & TIN & IND \\
English & @USER We are a country of morons & OFF & TIN & GRP \\
\bottomrule
\end{tabular}
}
\vspace{-2mm}
\caption{Annotated examples for all subtasks and languages.}
\label{T:examples}
\end{table*}

\paragraph{English} For English, we provided two datasets: OLID from OffensEval-2019~\cite{OLID}, and SOLID, which is a new dataset we created for the task~\cite{SOLID}.
SOLID is an abbreviation for Semi-Supervised Offensive Language Identification Dataset, and it contains 9,089,140 English tweets, which makes it the largest dataset of its kind. For SOLID, we collected random tweets using the 20 most common English stopwords such as \emph{the}, \emph{of}, \emph{and}, \emph{to}, etc. Then, we labeled the collected tweets in a semi-supervised manner using democratic co-training, with OLID as a seed dataset. For the co-training, we used four models with different inductive biases: PMI~\cite{turney2003measuring}, FastText~\cite{joulin2016bag}, LSTM~\cite{hochreiter1997long}, and BERT~\cite{devlin2019bert}. We selected the OFF tweets for the test set using this semi-supervised process and we then annotated them manually for all subtasks. We further added 2,500 NOT tweets using this process without further annotation. We computed a Fleiss' $\kappa$ Inter-Annotator Agreement (IAA) on a small subset of instances that were predicted to be OFF, and obtained 0.988 for Level A (almost perfect agreement), 0.818 for Level B (substantial agreement), and 0.630 for Level C (moderate agreement). The annotation for Level C was more challenging as it is 3-way and also as sometimes there could be different types of targets mentioned in the offensive tweet, but the annotators were forced to choose only one label.

\paragraph{Arabic} The Arabic dataset consists of 10,000 tweets collected in April--May 2019 using the Twitter API with the language filter set to Arabic: \texttt{lang:ar}. In order to increase the chance of having offensive content, only tweets with two or more vocative particles (\emph{yA} in Arabic) were considered for annotation; the vocative particle is used mainly to direct the speech to a person or to a group, and it is widely observed in offensive communications in almost all Arabic dialects. This yielded 20\% offensive tweets in the final dataset.
The tweets were manually annotated (for Level A only) by a native speaker familiar with several Arabic dialects. A random subsample of offensive and non-offensive tweets were doubly annotated and the Fleiss $\kappa$ IAA was found to be 0.92.
More details can be found in \cite{mubarak2020arabic}.

\paragraph{Danish} The Danish dataset consists of 3,600 comments drawn from Facebook, Reddit, and a local newspaper, Ekstra Bladet\footnote{\url{http://ekstrabladet.dk/}}. The selection of the comments was partially seeded using abusive terms gathered during a crowd-sourced lexicon compilation; in order to ensure sufficient data diversity, this seeding was limited to half the data only. 
The training data was not divided into distinct training/development splits, and participants were encouraged to perform cross-validation, as we wanted to avoid issues that fixed splits can cause~\cite{gorman2019we}. The annotation (for Level A only) was performed at the individual comment level by males aged 25-40.
A full description of the dataset and an accompanying data statement~\cite{bender2018data} can be found in ~\cite{sigurbergsson2020offensive}.

\paragraph{Greek} The Offensive Greek Twitter Dataset (OGTD) used in this task is a compilation of 10,287 tweets. These tweets were sampled using popular and trending hashtags, including television programs such as series, reality and entertainment shows, along with some politically related tweets. Another portion of the dataset was fetched using pejorative terms and ``you are'' as keywords. This particular strategy was adopted with the hypothesis that TV and politics would gather a handful of offensive posts, along with tweets containing vulgar language for further investigation. A team of volunteer annotators participated in the annotation process (for Level A only), with each tweet being judged by three annotators. In cases of disagreement, labels with majority agreement above 66\% were selected as the actual tweet labels. The IAA was 0.78 (using Fleiss' $\kappa$ coefficient).  A full description of the dataset collection and annotation is detailed in~\cite{pitenis2020}.

\paragraph{Turkish} The Turkish dataset consists of over 35,000 tweets sampled uniformly from the Twitter stream and filtered using a list of the most frequent words in Turkish, as identified by Twitter. The tweets were annotated by volunteers (for Level A only). Most tweets were annotated by a single annotator.
The Cohen's $\kappa$ IAA calculated on 5,000 doubly-annotated tweets was 0.761.
Note that we did not include any specific method for spotting offensive language, e.g.,~filtering by offensive words, or following usual targets of offensive language. As a result, the distribution closely resembles the actual offensive language use on Twitter, with more non-offensive tweets than offensive tweets. More details about the sampling and the annotation process can be found in~\cite{coltekin2020}.

\section{Task Participation}
\label{sec:taskparticipation}

A total of 528 teams signed up to participate in the task, and 145 of them submitted results: 6 teams made submissions for all five languages, 19 did so for four languages, 11 worked on three languages, 13 on two languages, and 96 focused on just one language.
Tables~\ref{table:team:participation:part1}, \ref{table:team:participation:part2}, and \ref{table:team:participation:part3} show a summary of which team participated in which task.
A total of 70 teams submitted system description papers, which are listed in Table~\ref{tab:teams}. Below, we analyze the representation and the models used for all language tracks.

\paragraph{Representation} The vast majority of teams used some kind of pre-trained embeddings such as contextualized Transformers~\cite{vaswani2017attention} and ELMo~\cite{peters2018deep} embeddings. The most popular Transformers were BERT~\cite{devlin2019bert}, RoBERTa~\cite{liu2019roberta}, and the multi-lingual mBERT~\cite{devlin2019bert}.\footnote{Note that there are some issues with the way mBERT processes some languages, e.g., there is no word segmentation for Arabic, the Danish {\aa}/aa mapping is not handled properly~\cite{stromberg2020danish}, etc.}

Many teams also used context-independent embeddings from word2vec~\cite{mikolov2013distributed} or GloVe~\cite{pennington2014glove}, including language-specific embeddings such as Mazajak~\cite{farha2019mazajak} for Arabic. Some teams used other techniques: word $n$-grams, character $n$-grams, lexicons for sentiment analysis, and lexicon of offensive words.
Other representations included emoji priors extracted from the weakly supervised SOLID dataset for English, and sentiment analysis using NLTK~\cite{bird2009natural}, Vader~\cite{hutto2014vader}, and FLAIR~\cite{akbik2018coling}.

\paragraph{Machine learning models} In terms of machine learning models, most teams used some kind of pre-trained Transformers: typically BERT, but RoBERTa, XLM-RoBERTa~\cite{conneau2019unsupervised}, ALBERT~\cite{lan2019albert}, and GPT-2~\cite{radford2019language} were also popular.
Other popular models included CNNs~\cite{fukushima1980neocognitron}, RNNs~\cite{rumelhart1986learning}, and GRUs~\cite{cho2014learning}. Older models such as SVMs~\cite{cortes1995support} were also used, typically as part of ensembles.

\section{English Track}
\label{sec:english}

A total of 87 teams made submissions for the English track (23 of them participated in the 2019 edition of the task): 27 teams participated in all three English subtasks, 18 teams participated in two English subtasks, and 42 focused on one English subtask only.

\paragraph{Pre-processing and normalization} Most teams performed some kind of pre-processing (67 teams) or text normalization (26 teams), which are typical steps when working with tweets. Text normalization included various text transformations such as converting emojis to plain text,\footnote{\url{http://github.com/carpedm20/emoji}} segmenting hashtags,\footnote{\url{http://github.com/grantjenks/python-wordsegment}} general tweet text normalization~\cite{satapathy2019seq2seq}, abbreviation expansion, bad word replacement, error correction, lowercasing, stemming, and/or lemmatization.
Other techniques included the removal of @user mentions, URLs, hashtags, emojis, emails, dates, numbers, punctuation, consecutive character repetitions, offensive words, and/or stop words.
 
\paragraph{Additional data} Most teams found the weakly supervised SOLID dataset useful, and 58 teams ended up using it in their systems. Another six teams gave it a try, but could not benefit from it, and the remaining teams only used the manually annotated training data. Some teams used additional datasets from HASOC-2019~\cite{hasoc2019}, the Kaggle competitions on Detecting Insults in Social Commentary\footnote{\url{http://www.kaggle.com/c/detecting-insults-in-social-commentary}} and Toxic Comment Classification\footnote{\url{http://www.kaggle.com/c/jigsaw-toxic-comment-classification-challenge}}, the TRAC-2018 shared task on Aggression Identification~\cite{trac2018report,KUMAR18.861}, the Wikipedia Detox dataset~\cite{wulczyn2017ex},
and the datasets from
\cite{hateoffensive} and \cite{wulczyn2017ex}, as well as some lexicons such as HurtLex~\cite{DBLP:conf/clic-it/BassignanaBP18} and Hatebase.\footnote{\url{http://hatebase.org/}} Finally, one team created their own dataset.
 
\subsection{Subtask A}

A total of 82 teams made submissions for subtask A, and the results can be seen in Table~\ref{tab:res_en_A}.
This was the most popular subtask among all subtasks and across all languages.
The best team UHH-LT achieved an F1 score of 0.9204 using an ensemble of ALBERT models of different sizes. 
The team ranked second was UHH-LT with an F1 score of 0.9204, and it used RoBERTa-large that was fine-tuned on the SOLID dataset in an unsupervised way, i.e.,~using the MLM objective.
The third team, Galileo, achieved an F1 score of 0.9198, using an ensemble that combined XLM-RoBERTa-base and XLM-RoBERTa-large trained on the subtask A data for all languages.
The top-10 teams used BERT, RoBERTa or XLM-RoBERTa, sometimes as part of ensembles that also included CNNs and LSTMs~\cite{hochreiter1997long}.
Overall, the competition for this subtask was very strong, and the scores are very close: the teams ranked 2--16 are within one point in the third decimal place, and those ranked 2--59 are within two absolute points in the second decimal place from the best team. All but one team beat the majority class baseline (we suspect that team might have accidentally flipped their predicted labels).

\begin{table*}[ht!]
\scalebox{0.88}{
\begin{tabular}{clc|clc|clc}
\toprule
\textbf{\#} & \textbf{Team} & \textbf{Score} & \textbf{\#} & \textbf{Team} & \textbf{Score} & \textbf{\#} & \textbf{Team} & \textbf{Score}\\ 
\midrule
 1 & UHH-LT & 0.9204  &  29 & UTFPR & 0.9094  &  57 & OffensSzeged & 0.9032 \\
 2 & Galileo & 0.9198  &  30 & IU-UM@LING & 0.9094  &  58 & FBK-DH & 0.9032 \\
 3 & Rouges & 0.9187  &  31 & TAC & 0.9093  &  59 & RGCL & 0.9006 \\
 4 & GUIR & 0.9166  &  32 & SSN\_NLP\_MLRG & 0.9092  &  60 & byteam & 0.8994 \\
 5 & KS@LTH & 0.9162  &  33 & Hitachi & 0.9091  &  61 & ANDES & 0.8990 \\
 6 & kungfupanda & 0.9151  &  34 & CoLi @ UdS & 0.9091  &  62 & PUM & 0.8973 \\
 7 & TysonYU & 0.9146  &  35 & XD & 0.9090  &  63 & NUIG & 0.8927 \\
 8 & AlexU-BackTranslation-TL & 0.9139  &  36 & UoB & 0.9090  &  64 & I2C & 0.8919 \\
 9 & SpurthiAH & 0.9136  &  37 & PAI-NLP & 0.9089  &  65 & sonal.kumari & 0.8900 \\
 10 & amsqr & 0.9135  &  38 & PingANPAI & 0.9089  &  66 & IJS & 0.8887 \\
 11 & m20170548 & 0.9134  &  39 & VerifiedXiaoPAI & 0.9089  &  67 & IR3218-UI & 0.8843 \\
 12 & Coffee\_Latte & 0.9132  &  40 & nlpUP & 0.9089  &  68 & TeamKGP & 0.8822 \\
 13 & wac81 & 0.9129  &  41 & NLP\_Passau & 0.9088  &  69 & UNT Linguistics & 0.8820 \\
 14 & NLPDove & 0.9129  &  42 & TheNorth & 0.9087  &  70 & janecek1 & 0.8744 \\
 15 & UJNLP & 0.9128  &  43 & problemConquero & 0.9085  &  71 & Team Oulu & 0.8655 \\
 16 & ARA & 0.9119  &  44 & Lee & 0.9084  &  72 & TECHSSN & 0.8655 \\
 17 & Ferryman & 0.9115  &  45 & Wu427 & 0.9081  &  73 & KDELAB & 0.8653 \\
 18 & ALT & 0.9114  &  46 & ITNLP & 0.9081  &  74 & HateLab & 0.8617 \\
 19 & SINAI & 0.9105  &  47 & Better Place & 0.9077  &  75 & IASBS & 0.8577 \\
 20 & MindCoders & 0.9105  &  48 & IIITG-ADBU & 0.9075  &  76 & IUST & 0.8288 \\
 21 & IRLab\_DAIICT & 0.9104  &  49 & doxaAI & 0.9075  &  77 & Duluth & 0.7714 \\
 22 & erfan & 0.9103  &  50 & NTU\_NLP & 0.9067  &  78 & RTNLU & 0.7665 \\
 23 & Light & 0.9103  &  51 & FERMI & 0.9065  &  79 & KarthikaS & 0.6351 \\
 24 & KAFK & 0.9099  &  52 & AdelaideCyC & 0.9063  &  80 & Bodensee & 0.4954 \\
 25 & PALI & 0.9098  &  53 & INGEOTEC & 0.9061  &   & \textbf{Majority Baseline} & 0.4193 \\
 26 & PRHLT-UPV& 0.9097  &  54 & PGSG & 0.9060  &  81 & IRlab@IITV & 0.0728 \\
 27 & YNU\_oxz & 0.9097  &  55 & SRIB2020 & 0.9048  &  & & \\
 28 & IITP-AINLPML & 0.9094  &  56 & GruPaTo & 0.9036  &  & & \\
 \bottomrule
\end{tabular}
}
\caption{Results for English subtask A, ordered by macro-averaged F1 in descending order.}
\label{tab:res_en_A}
\end{table*}

\subsection{Subtask B}

A total of 41 teams made submissions for subtask B, and the results can be seen in Table~\ref{tab:res_en_B}. The best team is Galileo (which were third on subtask A), whose ensemble model achieved an F1 score of 0.7462. The second-place team, PGSG, used a complex teacher-student architecture built on top of a BERT-LSTM model, which was fine-tuned on the SOLID dataset in an unsupervised way, i.e.,~optimizing for the MLM objective. NTU\_NLP was ranked third with an F1 score of 0.6906. They tackled subtasks A, B, and C as part of a multi-task BERT-based model. Overall, the differences in the scores for subtask B are much larger than for subtask A. For example, the 4th team is two points behind the third one and seven points behind the first one. The top-ranking teams used BERT-based Transformer models, and all but four teams could improve over the majority class baseline.

\begin{table*}[ht!]
\scalebox{0.88}{
\begin{tabular}{clc|clc|clc}
\toprule
\textbf{\#} & \textbf{Team} & \textbf{Score} & \textbf{\#} & \textbf{Team} & \textbf{Score} & \textbf{\#} & \textbf{Team} & \textbf{Score}\\ \midrule 
 1 & Galileo & 0.7462  &  15 & Wu427 & 0.6208  &  29 & PALI & 0.5533 \\
 2 & PGSG & 0.7362  &  16 & UNT Linguistics & 0.6174  &  30 & AdelaideCyC & 0.5524 \\
 3 & NTU\_NLP & 0.6906  &  17 & I2C & 0.6012  &  31 & KAFK & 0.5518 \\
 4 & UoB & 0.6734  &  18 & PRHLT-UPV & 0.5987  &  32 & PAI-NLP & 0.5451 \\
 5 & TysonYU & 0.6687  &  19 & SRIB2020 & 0.5805  &  33 & VerifiedXiaoPAI & 0.5451 \\
 6 & GUIR & 0.6650  &  20 & FERMI & 0.5804  &  34 & Duluth & 0.5382 \\
 7 & UHH-LT & 0.6598  &  21 & IU-UM@LING & 0.5746  &  35 & Bodensee & 0.4926 \\
 8 & Ferryman & 0.6576  &  22 & PingANPAI & 0.5687  &  36 & TECHSSN & 0.3894 \\
 9 & IIITG-ADBU & 0.6528  &  23 & nlpUP & 0.5687  &  37 & KarthikaS & 0.3741 \\
 10 & CoLi @ UdS & 0.6445  &  24 & Team Oulu & 0.5676  &   & \textbf{Majority Baseline} & 0.3741 \\
 11 & IRLab\_DAIICT & 0.6412  &  25 & KDELAB & 0.5638  &  38 & IRlab@IITV & 0.2950 \\
 12 & INGEOTEC & 0.6321  &  26 & wac81 & 0.5627  &  39 & SSN\_NLP\_MLRG & 0.2912 \\
 13 & HateLab & 0.6303  &  27 & IITP-AINLPML & 0.5569  &  40 & IJS & 0.2841 \\
 14 & AlexU-BackTranslation-TL & 0.6300  &  28 & problemConquero & 0.5569  &  41 & KEIS@JUST & 0.2777 \\
 \bottomrule
\end{tabular}
}
\caption{Results for English subtask B, ordered by macro-averaged F1 in descending order.}
\label{tab:res_en_B}
\end{table*}

\begin{table*}[ht!]
\scalebox{0.88}{
\begin{tabular}{clc|clc|clc}
\toprule
\textbf{\#} & \textbf{Team} & \textbf{Score} & \textbf{\#} & \textbf{Team} & \textbf{Score} & \textbf{\#} & \textbf{Team} & \textbf{Score}\\ \midrule 
 1 & Galileo & 0.7145  &  14 & KAFK & 0.6168  &  27 & nlpUP & 0.5515 \\
 2 & LT@Helsinki & 0.6700  &  15 & ssn\_nlp & 0.6116  &  28 & IS & 0.5355 \\
 3 & PRHLT-UPV & 0.6692  &  16 & IJS & 0.6094  &  29 & sonal.kumari & 0.5260 \\
 4 & UHH-LT & 0.6683  &  17 & PALI & 0.6015  &  30 & SRIB2020 & 0.5147 \\
 5 & ITNLP & 0.6543  &  18 & FERMI & 0.5882  &  31 & KEIS@JUST & 0.4817 \\
 6 & wac81 & 0.6489  &  19 & problemConquero & 0.5871  &  32 & ultraviolet & 0.4776 \\
 7 & PUM & 0.6473  &  20 & Ferryman & 0.5809  &  33 & HateLab & 0.4535 \\
 8 & PingANPAI & 0.6394  &  21 & AlexU-BackTranslation-TL & 0.5761  &  34 & Bodensee & 0.3462 \\
 9 & IITP-AINLPML & 0.6388  &  22 & IIITG-ADBU & 0.5756  &  35 & Team Oulu & 0.3220 \\
 10 & PAI-NLP & 0.6347  &  23 & Duluth & 0.5744  &  36 & SSN\_NLP\_MLRG & 0.3178 \\
 11 & GUIR & 0.6319  &  24 & KDELAB & 0.5720  &   & \textbf{Majority Baseline} & 0.2704 \\
 12 & IU-UM@LING & 0.6265  &  25 & NTU\_NLP & 0.5695  &  & & \\
 13 & AdelaideCyC & 0.6232  &  26 & INGEOTEC & 0.5626  &  & & \\
 \bottomrule
\end{tabular}
}
\caption{Results for English subtask C, ordered by macro-averaged F1 in descending order.}
\label{tab:res_en_C}
\end{table*}

\subsection{Subtask C}

A total of 37 teams made submissions for subtask C and the results are shown in Table~\ref{tab:res_en_C}.
The best team was once again Galileo, with an F1 score of 0.7145. LT@Helsinki was ranked second with an F1 score of 0.6700. They used fine-tuned BERT with oversampling to improve class imbalance. The third best system was PRHLT-UPV with an F1 score of 0.6692, which combines BERT with hand-crafted features; it is followed very closely by UHH-LT at rank 4, which achieved an F1 score of 0.6683. 
This subtask is also dominated by BERT-based models, and all teams outperformed the majority class baseline. 

Note that the absolute F1-scores obtained by the best teams in the English subtasks A and C are substantially higher than the scores obtained by the best teams in OffensEval-2019: 0.9223 vs. 0.8290 for subtask A and 0.7145 vs. 0.6600 for subtask C. This suggests that the much larger SOLID dataset made available in OffensEval-2020 helped the models make more accurate predictions. 

Furthermore, it suggests that the weakly supervised method used to compile and annotate SOLID is a viable alternative to popular purely manual annotation approaches. A more detailed analysis of the systems' performances will be carried out in order to determine the contribution of the SOLID dataset for the results.
 
\subsection{Best Systems}

We provide some more details about the approaches used by the top teams for each subtask. We use subindices to show their rank for each subtask. Additional summaries for some of the best teams can be found in Appendix~\ref{app:A}.

\paragraph{Galileo \scriptsize{\texttt{(A:3,B:1,C:1)}}} This team was ranked 3rd, 1st, and 1st on the English subtasks A, B, and C, respectively. This is also the only team ranked among the top-3 across all languages.
For subtask A, they used multi-lingual pre-trained Transformers based on XLM-RoBERTa, followed by multi-lingual fine-tuning using the OffensEval data. Ultimately, they submitted an ensemble that combined XLM-RoBERTa-base and XLM-RoBERTa-large, achieving an F1 score of 0.9198.
For subtasks B and C, they used knowledge distillation in a teacher-student framework, using Transformers such as ALBERT and ERNIE 2.0~\cite{sun2019ernie} as teacher models, achieving an F1 score of 0.7462 and 0.7145, for subtasks B and C respectively.

\paragraph{UHH-LT \scriptsize{\texttt{(A:1)}}} This team was ranked 1st on subtask A with an F1 score of 0.9223.
They fine-tuned different Transformer models on the OLID training data, and then combined them into an ensemble. They experimented with BERT-base and BERT-large (uncased), RoBERTa-base and RoBERTa-large, XLM-RoBERTa, and four different ALBERT models (large-v1, large-v2, xxlarge-v1, and xxlarge-v2). In their official submission, they used an ensemble combining different ALBERT models. They did not use the labels of the SOLID dataset, but found the tweets it contained nevertheless useful for unsupervised fine-tuning (i.e., using the MLM objective) of the pre-trained Transformers.


\section{Arabic Track}
\label{sec:arabic}

A total of 108 teams registered to participate in the Arabic track, and ultimately 53 teams entered the competition with at least one valid submission. Among them, ten teams participated in the Arabic track only, while the rest participated in other languages in addition to Arabic. This was the second shared task for Arabic after the one at the 4th workshop on Open-Source Arabic Corpora and Processing Tools~\cite{mubarak-EtAl:2020:OSACT}, which had different settings and less participating teams.

\begin{table*}[t!]
\centering
\scalebox{0.88}{
\begin{tabular}{clc|clc|clc}
\toprule
\textbf{\#} & \textbf{Team} & \textbf{Score} & \textbf{\#} & \textbf{Team} & \textbf{Score} & \textbf{\#} & \textbf{Team} & \textbf{Score}\\ \midrule 
 1 & ALAMIHamza & 0.9017 & 21 & SaiSakethAluru & 0.8455 & 41 & tharindu & 0.7881 \\ 
2 & ALT & 0.9016 & 22 & will\_go & 0.8440 & 42 & PRHLT-UPV & 0.7868 \\ 
3 & Galileo & 0.8989 & 23 & erfan & 0.8418 & 43 & IRlab@IITV & 0.7793 \\ 
4 & KUISAIL & 0.8972 & 24 & ANDES & 0.8402 & 44 & yemen2016 & 0.7721 \\ 
5 & AMR-KELEG & 0.8958 & 25 & Bushr & 0.8395 & 45 & saroarj & 0.7474 \\ 
6 & KS@LTH & 0.8902 & 26 & klaralang & 0.8241 & 46 & kxkajava & 0.7306 \\ 
7 & iaf7 & 0.8778 & 27 & zoher\_orabe & 0.8221 & 47 & frankakorpel & 0.7251 \\ 
8 & INGEOTEC & 0.8744 & 28 & mircea.tanase & 0.8220 & 48 & COMA & 0.5436 \\ 
9 & BhamNLP & 0.8714 & 29 & machouz & 0.8216 & 49 & JCT & 0.4959 \\ 
10 & yasserotiefy & 0.8691 & 30 & orabia & 0.8198 & 50 & FBK-DH & 0.4642 \\ 
11 & SAJA & 0.8655 & 31 & Taha & 0.8183 & 51 & sonal.kumari & 0.4536 \\ 
12 & Ferryman & 0.8592 & 32 & hamadanayel & 0.8182 & 52 & CyberTronics & 0.4466 \\ 
13 & SAFA & 0.8555 & 33 & CoLi @ UdS & 0.8176 & 53 & SpurthiAH & 0.4451 \\ 
14 & hhaddad & 0.8520 & 34 & fatemah & 0.8147 &  &  \textbf{Majority Baseline} & 0.4441 \\ 
15 & TAC & 0.8519 & 35 & jbern & 0.8125 &  &  &  \\ 
16 & saradhix & 0.8500 & 36 & zahra.raj & 0.8057 &  &  &  \\ 
17 & lukez & 0.8498 & 37 & I2C & 0.8056 &  &  &  \\ 
18 & Rouges & 0.8480 & 38 & jlee24282 & 0.8024 &  &  &  \\ 
19 & TysonYU & 0.8474 & 39 & problemConquero & 0.8021 &  &  &  \\ 
20 & NLPDove & 0.8455 & 40 & asking28 & 0.8002 &  &  &  \\ \bottomrule
\end{tabular}
}
\caption{Results for Arabic subtask A, ordered by macro-averaged F1 in descending order.}
\label{tab:res_ar_A}
\end{table*}

\paragraph{Pre-processing and normalization} Most teams performed some kind of pre-processing or text normalization, e.g.,~Hamza shapes, Alif Maqsoura, Taa Marbouta, diacritics, non-Arabic characters, etc., and only one team replaced emojis with their textual counter-parts.

\subsection{Results}

Table~\ref{tab:res_ar_A} shows the teams and the F1 scores they achieved for the Arabic subtask A. The majority class baseline had an F1 score of 0.4441, and several teams achieved results that doubled that baseline score. The best-performing team was ALAMIHamza with an F1 score of 0.9017. The second-best team, ALT, was almost tied with the winner, with an F1 score of 0.9016. The Galileo team was third with an F1 score of 0.8989.
A summary of the approaches taken by the top-performing teams can be found in Appendix~\ref{app:A}; here we briefly describe the winning system:

\paragraph{ALAMIHamza\scriptsize{\texttt{(A:1)}}}
The winning team achieved the highest F1-score using BERT to encode Arabic tweets, followed by a sigmoid classifier. They further performed translation of the meaning of emojis.


\section{Danish Track}
\label{sec:danish}

A total of 72 teams registered to participate in the Danish track, and 39 of them actually made official submissions on the test dataset. This is the first shared task on offensive language identification to include Danish, and the dataset provided to the OffensEval-2020 participants is an extended version of the one from \cite{sigurbergsson2020offensive}.

\begin{table*}[t!]
\centering
\scalebox{0.88}{
\begin{tabular}{clc|clc|clc}
\toprule
\textbf{\#} & \textbf{Team} & \textbf{Score} & \textbf{\#} & \textbf{Team} & \textbf{Score} & \textbf{\#} & \textbf{Team} & \textbf{Score}\\ \midrule 
1	& LT@Helsinki	& 0.8119	& 14	& Rouges	& 0.7587	& 27	& TeamKGP	& 0.6973\\
2	& Galileo	& 0.8021	& 14	& Smatgrisene	& 0.7587	& 28	& Stormbreaker	& 0.6842\\
3	& NLPDove	& 0.7923	& 16	& machouz	& 0.7561	& 29	& TAC	& 0.6819\\
4	& FBK-DH	& 0.7766	& 17	& IU-UM@LING	& 0.7553	& 30	& Sonal	& 0.6711\\
5	& KS@LTH	& 0.7750	& 18	& Ferryman	& 0.7525	& 31	& RGCL	& 0.6556\\
6	& JCT	& 0.7741	    & 19	& MindCoders	& 0.7380	& 32	& PRHLT-UPV	& 0.6369\\
7	& ANDES	& 0.7723	& 20	& ARA	& 0.7267	& 33	& IUST	& 0.6226\\
8	& TysonYU	& 0.7685	& 21	& INGEOTEC	& 0.7237	& 34	& SRIB2020	& 0.6127\\
8	& FERMI	& 0.7685	    & 22	& KUISAIL	& 0.7231	& 35	& IR3218-UI	& 0.5736\\
10	& NLP\_Passau	& 0.7673	& 23	& JAK	& 0.7086	& 36	& SSN\_NLP\_MLRG	& 0.5678\\
11	& GruPaTo	& 0.7620	& 24	& LIIR	& 0.7019	& 37	& Team Oulu	& 0.5587\\
12	& KEIS@JUST	& 0.7612	& 25	& MeisterMorxrc	& 0.6998 & 38	& IJS	& 0.4913 \\
13	& will\_go	& 0.7596	& 26	& problemConquero	& 0.6974 &	& \bf Majority Baseline	& 0.4668 \\
\bottomrule
\end{tabular}}
\caption{Results for Danish subtask A, ordered by macro-averaged F1 in descending order.}
\label{tab:res_da_A}
\end{table*}


\paragraph{Pre-processing and normalization} Many teams used the pre-processing included in the relevant embedding model, e.g.,~BPE~\cite{heinzerling2018bpemb} and WordPiece. Other pre-processing techniques included emoji normalization, spelling correction, sentiment tagging, lexical and regex-based term and phrase flagging, and hashtag segmentation.

\subsection{Results}
The results are shown in Table~\ref{tab:res_da_A}. We can see that all teams managed to outperform the majority class baseline. Moreover, all but one team improved over a FastText baseline (F1 = 0.5148), and most teams achieved an F1 score of 0.7 or higher. Interestingly, one of the top-ranked teams, JCT, was entirely non-neural.
 
\paragraph{LT@Helsinki \scriptsize{\texttt{(A:1)}}} The winning team LT@Helsinki used NordicBERT for representation, as provided by BotXO.\footnote{See \url{http://github.com/botxo/nordic_bert}} NordicBERT is customized to Danish, and avoids some of the pre-processing noise and ambiguity introduced by other popular BERT implementations. The team further reduced orthographic lengthening to maximum two repeated characters, converted emojis to sentiment scores, and used co-occurrences of hashtags and references to usernames. They tuned the hyper-parameters of their model using 10-fold cross validation.


\section{Greek Track}\label{sec:greek}

A total of 71 teams registered to participate in the Greek track, and ultimately 37 of them made an official submission on the test dataset. This is the first shared task on offensive language identification to include Greek, and the dataset provided to the OffensEval-2020 participants is an extended version of the one from \cite{pitenis2020}.

\begin{table*}[ht!]
\centering 
\scalebox{0.88}{
\begin{tabular}{clc|clc|clc}
\toprule
\textbf{\#} & \textbf{Team} & \textbf{Score} & \textbf{\#} & \textbf{Team} & \textbf{Score} & \textbf{\#} & \textbf{Team} & \textbf{Score}\\ \midrule 
 1 & NLPDove & 0.8522   &  14 & CoLi @ UdS & 0.8147  &  27 & IUST & 0.7756 \\ 
 2 & Galileo & 0.8507  &  15 & TAC & 0.8141  &  28 & KEIS@JUST & 0.7730  \\ 
 3 & KS@LTH & 0.8481  &  16 & IU-UM@LING & 0.8140  &  29 & FBK-DH & 0.7700 \\ 
 4 & KUISAIL & 0.8432 &  17 & MindCoders & 0.8137  &  30 & Team Oulu & 0.7615 \\ 
 5 & IJS & 0.8329  &  18 & RGCL & 0.8135  &  31 & JCT & 0.7568 \\ 
 6 & SU-NLP & 0.8317  &  19 & problemConquero & 0.8115  &  32 & IRlab@IITV & 0.7181 \\ 
 7 & LT@Helsinki & 0.8258  &  20 & Rouges & 0.8030  &  33 & TeamKGP & 0.7041 \\ 
 8 & FERMI & 0.8231   &  21 & TysonYU & 0.8022  &  34 & SSN\_NLP\_MLRG & 0.6779 \\ 
 9 & Ferryman & 0.8222  &  22 & Sonal & 0.8017  &  35 & fatemah & 0.6036 \\ 
 10 & INGEOTEC & 0.8197 &  23 & JAK & 0.7956  &  36 & CyberTronics & 0.4265 \\
 11 & will\_go & 0.8176  &  24 & ARA & 0.7828  &  & {\bf Majority Baseline} & 0.4202 \\ 
 12 & ANDES & 0.8153  &  25 & machouz & 0.7820 & 37 & Stormbreaker & 0.2688 \\ 
 13 & LIIR & 0.8148  &  26 & PRHLT-UPV & 0.7763  &  & & \\ \bottomrule
\end{tabular}
}
\caption{Results for Greek subtask A, ordered by macro-averaged F1 in descending order.}
\label{tab:res_el_A}
\end{table*}

\paragraph{Pre-processing and normalization} The participants experimented with various pre-processing and text normalization techniques, similarly to what was done for the other languages above. One team further reported replacement of emojis with their textual equivalent.

\subsection{Results}

The evaluation results are shown in Table~\ref{tab:res_el_A}. The top team, NLPDove, achieved an F1 score of 0.852, with Galileo coming close at the second place with an F1 score of 0.851. The KS@LTH team was ranked third with an F1 score of 0.848.  It is no surprise that the majority of the high-ranking submissions and participants used large-scale pre-trained Transformers, with BERT being the most prominent among them, along with wordwvec-style non-contextualized pre-trained word embeddings.

\paragraph{NLPDove \scriptsize{\texttt{(A:1)}}}
The winning team NLPDove used pre-trained word embeddings from mBERT, which they fine-tuned using the training data. A domain-specific vocabulary was generated by running the WordPiece algorithm~\cite{schuster2012japanese} and using embeddings for extended vocabulary to pre-train and fine-tune the model.

\section{Turkish Track}\label{sect:turkish}

A total of 86 teams registered to participate in the Turkish track, and ultimately 46 of them made an official submission on the test dataset. All teams except for one participated in at least one other track.  This is the first shared task on offensive language identification to include Turkish, and the dataset provided to the OffensEval-2020 participants is an extended version of the one from \cite{coltekin2020}.

\pgfplotstableread[col sep=comma,trim cells]{turkish-leaderboard-detailed.csv}{\trresults}
\pgfplotstablegetrowsof{\trresults}
\let\trnumberofrows=\pgfplotsretval
\pgfplotstablegetelem{0}{{f1-macro}}\of{\trresults}
\let\firstF=\pgfplotsretval
\pgfplotstablegetelem{1}{{f1-macro}}\of{\trresults}
\let\secondF=\pgfplotsretval
\pgfplotstablegetelem{2}{{f1-macro}}\of{\trresults}
\let\thirdF=\pgfplotsretval

\begin{table}[ht!]
  \centering
  \scalebox{0.88}{
    \pgfplotstabletypeset[%
      every head row/.style={before row=\toprule,
        after row={\midrule},
      },
      every last row/.style={ after row=\bottomrule},
      columns={rank,team,{f1-macro},rank,team,{f1-macro},rank,team,{f1-macro}},
      display columns/0/.style={select equal part entry of={0}{3},
        column type/.add={}{@{\hspace{1em}}}},
      display columns/1/.style={select equal part entry of={0}{3}},
      display columns/2/.style={select equal part entry of={0}{3},
        column type/.add={}{|}},
      display columns/3/.style={select equal part entry of={1}{3},
        column type/.add={}{@{\hspace{1em}}}},
      display columns/4/.style={select equal part entry of={1}{3}},
      display columns/5/.style={select equal part entry of={1}{3},
        column type/.add={}{|}},
      display columns/6/.style={select equal part entry of={2}{3},
        column type/.add={}{@{\hspace{1em}}}},
      display columns/7/.style={select equal part entry of={2}{3}},
      display columns/8/.style={select equal part entry of={2}{3}},
      columns/{f1-macro}/.style={
        column name={\textbf{Score}},
        fixed,fixed zerofill,precision=4,
        column type=r,
      },
      columns/rank/.style={
        column name={\textbf{\#}},
        fixed,precision=0,
        column type=r,
      },
      columns/team/.style={
        column name={\textbf{Team}},
        column type=l,
        string type,
        postproc cell content/.code={%
           \pgfplotsutilstrreplace{_}{\_}{##1}%
           \pgfkeyslet{/pgfplots/table/@cell content}\pgfplotsretval
        },
      }
    ]{\trresults}
  }
  \caption{\label{tab:res_tr}Results for Turkish subtask A, ordered by macro-averaged F1 in descending order.}
\end{table}

\subsection{Results}

The results are shown in Table~\ref{tab:res_tr}. We can see that team Galileo achieved the highest macro-averaged F1 score of 0.8258, followed by SU-NLP and KUI-SAIL with F1 scores of 0.8167 and 0.8141, respectively. Note that the latter two teams are from Turkey, and they used some language-specific resources and tuning. Most results were in the interval 0.7--0.8, and almost all teams managed to outperform the majority class baseline, which had an F1 score of 0.4435.

\paragraph{Galileo \scriptsize{\texttt{(A:1)}}}
The best team in the Turkish subtask A was Galileo, which achieved top results in several other tracks. Unlike the systems ranked second and third, Galileo's system is language-agnostic, and it used data for all five languages in a multi-lingual training setup.

\section{Conclusion and Future Work}
\label{sec:conclusion}

We presented the results of OffensEval-2020, which featured datasets in five languages: Arabic, Danish, English, Greek, and Turkish. For English, we had three subtasks, representing the three levels of the OLID hierarchy. For the other four languages, we had a subtask for the top-level of the OLID hierarchy only. A total of 528 teams signed up to participate in OffensEval-2020, and 145 of them actually submitted results across all languages and subtasks.

Out of the 145 participating teams, 96 teams participated in one language only, 13 teams participated in two languages, 11 in three languages, 19 in four languages, and 6 teams submitted systems for all five languages. The official submissions per language ranged from 37 (for Greek) to 81 (for English).  Finally, 70 of the 145 participating teams submitted system description papers, which is an all-time record.

The wide participation in the task allowed us to compare a number of approaches across different languages and datasets. Similarly to OffensEval-2019, we observed that the best systems for all languages and subtasks used large-scale BERT-style pre-trained Transformers such as BERT, RoBERTa, and mBERT. Unlike 2019, however, the multi-lingual nature of this year's data enabled cross-language approaches, which proved quite effective and were used by some of the top-ranked systems.

In future work, we plan to extend the task in several ways. First, we want to offer subtasks B and C for all five languages from OffensEval-2020. We further plan to add some additional languages, especially under-represented ones. Other interesting aspects to explore are code-mixing, e.g.,~mixing Arabic script and Latin alphabet in the same Arabic message, and code-switching, e.g.,~mixing Arabic and English words and phrases in the same message. Last but not least, we plan to cover a wider variety of social media platforms.

\section*{Acknowledgements}
This research was partly supported by the IT University of Copenhagen's Abusive Language Detection project. It is also supported by the Tanbih project at the Qatar Computing Research Institute, HBKU, which aims to limit the effect of ``fake news,'' propaganda and media bias by making users aware of what they are reading.

\bibliographystyle{coling}
\bibliography{offenseval2020,offenseval2020-system-papers}

\newpage

\appendix

\section{Best-Performing Teams}
~\label{app:A}

Below we present a short overview of the top-3 systems for all subtasks and for all languages:

\paragraph{Galileo \scriptsize{\texttt{(EN B:1, EN C:1, TR A:1; DK A:2, GR A:2; AR A:3, EN A:3)}}} This team was ranked 3rd, 1st, and 1st on the English subtasks A, B, and C, respectively; it was also ranked 1st for Turkish, 2nd for Greek and 3rd for Arabic and Danish. This is the only team ranked among the top-3 across all languages.
For subtask A (all languages), they used multi-lingual pre-trained Transformers based on XLM-RoBERTa, followed by multi-lingual fine-tuning using the OffensEval data. Ultimately, they submitted an ensemble that combined XLM-RoBERTa-base and XLM-RoBERTa-large.
For the English subtasks B and C, they used knowledge distillation in a teacher-student framework, using Transformers such as ALBERT and ERNIE 2.0~\cite{sun2019ernie} as teacher models.

\paragraph{UHH-LT \scriptsize{\texttt{(EN A:1)}}} This team was ranked 1st on the English subtask A. They fine-tuned different Transformer models on the OLID training data, and then combined them into an ensemble. They experimented with BERT-base and BERT-large (uncased), RoBERTa-base and RoBERTa-large, XLM-RoBERTa, and four different ALBERT models (large-v1, large-v2, xxlarge-v1, and xxlarge-v2). In their official submission, they used an ensemble combining different ALBERT models. They did not use the labels of the SOLID dataset, but found the tweets it contained nevertheless useful for unsupervised fine-tuning (i.e.,~using the MLM objective) of the pre-trained Transformers.

\paragraph{LT@Helsinki \scriptsize{\texttt{(DK A:1; EN C:2)}}} This team was ranked 1st for Danish and 2nd for English subtask C. For Danish, they used NordicBERT, which is customized to Danish, and avoids some of the pre-processing noise and ambiguity introduced by other popular BERT implementations. The team further reduced orthographic lengthening to maximum two repeated characters, converted emojis to sentiment scores, and used co-occurrences of hashtags and references to usernames. They tuned the hyper-parameters of their model using 10-fold cross validation. For English subtask C, they used a very simple approach: over-sample the training data to overcome the class imbalance, and then fine-tune BERT-base-uncased.

\paragraph{NLPDove \scriptsize{\texttt{(GR A:1; DK A:3)}}}
This team was ranked 1st for Greek and 3rd for Danish. This team used extensive preprocessing and two data augmentation strategies: using additional semi-supervised labels from SOLID with different thresholds, and cross-lingual transfer with data selection. They further proposed and used a new metric, Translation Embedding Distance, in order to measure the transferability of instances for cross-lingual data selection. Moreover, they used data from different languages to fine-tune an mBERT model. Ultimately, they used a majority vote ensemble of several mBERT models, with minor variations in the parameters.

\paragraph{ALAMIHamza\scriptsize{\texttt{(AR A:1)}}}
This team was ranked 1st for Arabic. They used BERT to encode Arabic tweets, followed by a sigmoid classifier. They further performed translation of the meaning of emojis.

\paragraph{PGSG \scriptsize{\texttt{(EN B:2)}}} The team was ranked 2nd on the English subtask B.
They first fine-tuned the BERT-Large, Uncased (Whole Word Masking) checkpoint using the tweets from SOLID, but ignoring their labels. For this, they optimized for the MLM objective only, without the Next Sentence Prediction loss in BERT.
Then, they built a BERT-LSTM model using this fine-tuned BERT, and adding LSTM layers on top of it, together with the [CLS] token. Finally, they used this architecture to train a Noisy Student model using the SOLID data.

\paragraph{ALT \scriptsize{\texttt{(AR A:2)}}}
The team was ranked 2nd for Arabic. They used an ensemble of SVM, CNN-BiLSTM and Multilingual BERT. The SVMs used character $n$-grams, word $n$-grams, word embeddings as features, while the CNN-BiLSTM learned character embeddings and further used pre-trained word embeddings as input.

\paragraph{SU-NLP \scriptsize{\texttt{(TR A:2)}}}
The team was ranked 2nd for Turkish. They used an ensemble of three different models: CNN-LSTM, BiLSTM-Attention, and BERT. They further used word embeddings, pre-trained on tweets, and BERTurk, BERT model for Turkish.

\paragraph{Rouges \scriptsize{\texttt{(EN A:3)}}}
The team was ranked 3rd for the English subtask A. They used XLM-RoBERTa fine-tuned sequentially on all languages in a particular order: English, then Turkish, then Greek, then Arabic, then Danish.

\paragraph{NTU\_NLP \scriptsize{\texttt{(EN B:3)}}} This team was ranked 3rd on the English subtask B.
They proposed a hierarchical multi-task learning approach that solves subtasks A, B, and C simultaneously, following the hierarchical structure of the annotation schema of the OLID dataset. Their architecture has three layers. The input of the first layer is the output of BERT, and its output (D1-OUT) is directly connected to the output layer for subtask A. The second layer's input is the BERT output concatenated with D1-OUT, and its output (D2-OUT) is directly connected to the output layer for subtask B. The third layer's input is the BERT output concatenated with D2-OUT, and its output is directly connected to the output layer for subtask C.

\paragraph{PRHLT-UPV \scriptsize{\texttt{(EN C:3)}}}
The team was ranked 3rd on the English subtask C. They used a combination of BERT and hand-crafted features, which were concatenated to the [CLS] representation from BERT. The features include the length of the tweets, the number of misspelled words, and the use of punctuation marks, emoticons, and noun phrases.

\paragraph{KS@LTH \scriptsize{\texttt{(GR: A:3)}}}
This team was ranked 3rd for Greek. They experimented with monolingual and cross-lingual models, BERT and XLM-RoBERTa model, respectively, and they found BERT to perform slightly better.

\paragraph{KUISAIL \scriptsize{\texttt{(TR: A:3)}}}
This team was ranked 3rd for Turkish. They combined BERTurk with a CNN, in a BERT-CNN model.

\newpage

\section{Participants}

\begin{table}[ht!]
\centering
\scalebox{0.75}{
\begin{tabular}{@{}ll|ll@{}}
\toprule
   {\bf Team}  & {\bf System Description Paper} & {\bf Team}  & {\bf System Description Paper}\\ 
\midrule
AdelaideCyC & \cite{offenseval2020ID45} & LISAC FSDM-USMBA & \cite{offenseval2020ID145} \\
AlexU-BackTranslation-TL & \cite{offenseval2020ID92} & LT@Helsinki & \cite{offenseval2020ID14} \\
ALT & \cite{offenseval2020ID87} & NAYEL & \cite{offenseval2020ID253} \\
amsqr & \cite{offenseval2020ID54} & NLP\_Passau & \cite{offenseval2020ID111} \\
ANDES & \cite{offenseval2020ID291} & NLPDove & \cite{offenseval2020ID287} \\
BhamNLP & \cite{offenseval2020ID224} & nlpUP & \cite{offenseval2020ID141} \\
JCT & \cite{offenseval2020ID186} & Nova-Wang & \cite{offenseval2020ID286} \\
BRUMS & \cite{offenseval2020ID188} & NTU\_NLP & \cite{offenseval2020ID178} \\
CoLi @ UdS & \cite{offenseval2020ID261} & NUIG & \cite{offenseval2020ID89} \\
CyberTronics & \cite{offenseval2020ID247} & Oulu & \cite{offenseval2020ID165} \\
DoTheMath & \cite{offenseval2020ID219} & PGSG & \cite{offenseval2020ID42} \\
Duluth & \cite{offenseval2020ID34} & pin\_cod\_ & \cite{offenseval2020ID22} \\
FBK-DH & \cite{offenseval2020ID44} & PRHLT-UPV & \cite{offenseval2020ID64} \\
Ferryman & \cite{offenseval2020ID168} & problemConquero & \cite{offenseval2020ID98} \\
Galileo & \cite{offenseval2020ID228} & PUM & \cite{offenseval2020ID273} \\
Garain & \cite{offenseval2020ID27} & Rouges & \cite{offenseval2020ID32} \\
GruPaTo & \cite{offenseval2020ID23} & SalamNET & \cite{offenseval2020ID295} \\
GUIR & \cite{offenseval2020ID266} & SINAI & \cite{offenseval2020ID86} \\
Hitachi & \cite{offenseval2020ID112} & Smatgrisene & \cite{offenseval2020ID184} \\
I2C & \cite{offenseval2020ID200} & Sonal.kumari & \cite{offenseval2020ID194} \\
iCompass & \cite{offenseval2020ID75} & SRIB2020 & \cite{offenseval2020:SRIB} \\
IIITG-ADBU & \cite{offenseval2020ID254} & SSN\_NLP\_MLRG & \cite{offenseval2020ID149} \\
IITP-AINLPML & \cite{offenseval2020ID239} & SU-NLP & \cite{offenseval2020ID284} \\
INGEOTEC & \cite{offenseval2020ID223} & TAC & \cite{offenseval2020ID79} \\
IR3218-UI & \cite{offenseval2020ID189} & TECHSSN & \cite{offenseval2020ID126} \\
IRlab@IITV & \cite{offenseval2020ID210} & TheNorth & \cite{offenseval2020ID279} \\
IRLab\_DAIICT & \cite{offenseval2020ID244} &  UHH-LT & \cite{offenseval2020ID20} \\
KAFK & \cite{offenseval2020ID192} & UJNLP & \cite{offenseval2020ID134} \\
KDELAB & \cite{offenseval2020ID262} &  UNT & \cite{offenseval2020ID204} \\
KEIS@JUST & \cite{offenseval2020ID217} & UoB & \cite{offenseval2020ID29}  \\
KS@LTH & \cite{offenseval2020ID28} & UPB & \cite{offenseval2020ID271} \\
KUISAIL & \cite{offenseval2020ID311} & UTFPR & \cite{offenseval2020ID48} \\
Kungfupanda & \cite{offenseval2020ID56} & WOLI & \cite{offenseval2020ID225} \\
Lee & \cite{offenseval2020ID175} & XD & \cite{offenseval2020ID303} \\
LIIR & \cite{offenseval2020ID10} & YNU\_oxz & \cite{offenseval2020ID230} \\
\bottomrule
\end{tabular}
}
\caption{The teams that participated in OffensEval-2020 and submitted system description papers and the corresponding reference to their papers.}
\label{tab:teams}
\end{table}

\begin{table}[tbh]
\small
\begin{tabular}{lccccccc}
\toprule
\bf Team & \bf A-Arabic & \bf A-Danish & \bf A-Greek & \bf A-Turkish & \bf A-English & \bf B-English & \bf C-English \\
\midrule
AdelaideCyC & & & & & & \faCheck & \\
AlexU-BackTranslation-TL & & & & & \faCheck & \faCheck & \faCheck \\
ALT & \faCheck & & & & \faCheck & & \\
alaeddin & & & & \faCheck & & & \\
ALAMIHamza & \faCheck & & & & & & \\
AMR-KELEG & \faCheck & & & & & & \\
amsqr & & & & & \faCheck & & \\
ANDES & \faCheck & \faCheck & \faCheck & \faCheck  & \faCheck & & \\
ARA & & \faCheck & \faCheck & \faCheck & \faCheck & & \\
asking28 & \faCheck & & & & & & \\
Better Place & & & & & \faCheck & & \\
BhamNLP & \faCheck & & & & & & \\
Bodensee & & & & & \faCheck & \faCheck & \faCheck \\
Bushr & \faCheck & & & & & & \\
byteam & & & & & \faCheck & & \\
Coffee\_Latte & & & & & \faCheck & & \\
CoLi @ UdS & \faCheck & & \faCheck & \faCheck & \faCheck & \faCheck & \\
COMA & \faCheck & & & & & & \\
CyberTronics & \faCheck & & \faCheck & \faCheck & & & \\
doxaAI & & & & & \faCheck & & \\
Duluth & & & & & \faCheck & \faCheck & \faCheck \\
erfan & \faCheck & & & & \faCheck & & \\
f\_shahaby & & & & \faCheck & & & \\
fatemah & \faCheck & & \faCheck & \faCheck & & & \\
FBK-DH & \faCheck & \faCheck & \faCheck & \faCheck & \faCheck & & \\
FERMI & & \faCheck & \faCheck & \faCheck & \faCheck & \faCheck & \faCheck \\
Ferryman & \faCheck & \faCheck & \faCheck & \faCheck & \faCheck & \faCheck & \faCheck \\
frankakorpel & \faCheck & & & & & & \\
Galileo & \faCheck & \faCheck & \faCheck & \faCheck & \faCheck & \faCheck & \faCheck \\
GruPaTo & & \faCheck & & \faCheck & \faCheck & & \\
GUIR & & & & & \faCheck & \faCheck & \faCheck \\
hamadanayel & \faCheck & & & & & & \\
HateLab & & & & & \faCheck & \faCheck & \faCheck \\
hhaddad & \faCheck & & & & & & \\
Hitachi & & & & & \faCheck & & \\
I2C & \faCheck & & & \faCheck & \faCheck & \faCheck & \\
iaf7 & \faCheck & & & & & & \\
IASBS & & & & \faCheck & \faCheck & & \\
IIITG-ADBU & & & & & \faCheck & \faCheck & \faCheck \\
IITP-AINLPML & & & & & \faCheck & \faCheck & \faCheck \\
IJS & & \faCheck & \faCheck & \faCheck & \faCheck & \faCheck & \faCheck \\
INGEOTEC & \faCheck & \faCheck & \faCheck & \faCheck & \faCheck & \faCheck & \faCheck \\
IR3218-UI & & \faCheck & & & \faCheck & & \\
IRlab@IITV & & & \faCheck & & \faCheck & \faCheck & \\
IRLab\_DAIICT & & & & & \faCheck & \faCheck & \\
IS & & & & & & & \faCheck \\
ITNLP & & & & & \faCheck & & \faCheck \\
IU-UM@LING & & \faCheck & \faCheck & \faCheck & \faCheck & \faCheck & \faCheck \\
IUST & & \faCheck & \faCheck & \faCheck & \faCheck & & \\
JAK & & \faCheck & \faCheck & \faCheck & & & \\
janecek1 & & & & & \faCheck & & \\
jbern & \faCheck & & & & & & \\
JCT & \faCheck & \faCheck & \faCheck & \faCheck & & & \\
jlee24282 & \faCheck & & & & & & \\
jooyeon Lee & & & & \faCheck & & & \\
KAFK & & & & & \faCheck & \faCheck & \faCheck \\
KarthikaS & & & & & \faCheck & \faCheck & \\
KDELAB & & & & & \faCheck & \faCheck & \faCheck \\
KEIS@JUST & & \faCheck & \faCheck & \faCheck & & \faCheck & \faCheck \\
\bottomrule
\end{tabular}
\caption{Overview of team participation in the subtasks (part 1).}
\label{table:team:participation:part1}
\end{table}

\begin{table}[tbh]
\small
\begin{tabular}{lccccccc}
\toprule
\bf Team & \bf A-Arabic & \bf A-Danish & \bf A-Greek & \bf A-Turkish & \bf A-English & \bf B-English & \bf C-English \\
\midrule
klaralang & \faCheck & & & & & & \\
KS@LTH & \faCheck & \faCheck & \faCheck & \faCheck & \faCheck & & \\
KUISAIL & \faCheck & & & \faCheck & & & \\
kungfupanda & & & & & \faCheck & & \\
kxkajava & \faCheck & & & & & & \\
Lee & & & & & \faCheck & & \\
Light & & & & & \faCheck & & \\
LIIR & & \faCheck & \faCheck & \faCheck & & & \\
LT@Helsinki & & \faCheck & \faCheck & \faCheck & & & \faCheck \\
lukez & \faCheck & & & & & & \\
m20170548 & & & & & \faCheck & & \\
machouz & \faCheck & \faCheck & \faCheck & \faCheck & & & \\
MeisterMorxrc & & \faCheck & & \faCheck & & & \\
MindCoders & & \faCheck & \faCheck & \faCheck & \faCheck & & \\
mircea.tanase & \faCheck & & & & & & \\
NLP\_Passau & & \faCheck & & \faCheck & \faCheck & & \\
NLPDove & \faCheck & \faCheck & \faCheck & \faCheck & \faCheck & & \\
nlpUP & & & & & \faCheck & \faCheck & \faCheck \\
NTU\_NLP & & & & & \faCheck & \faCheck & \faCheck \\
NUIG & & & & & \faCheck & & \\
OffensSzeged & & & & & \faCheck & & \\
orabia & \faCheck & & & & & & \\
Oulu & & & & \faCheck & & & \\
PAI-NLP & & & & & \faCheck & \faCheck & \faCheck \\
PALI & & & & & \faCheck & \faCheck & \faCheck \\
PGSG & & & & & \faCheck & \faCheck & \\
pin\_cod\_ & & & & \faCheck & & & \\
PingANPAI & & & & & \faCheck & \faCheck & \faCheck \\
PRHLT-UPV & \faCheck & \faCheck & \faCheck & \faCheck & \faCheck & \faCheck & \faCheck \\
problemConquero & \faCheck & \faCheck & \faCheck & \faCheck & \faCheck & \faCheck & \faCheck \\
PUM & & & & & \faCheck & & \faCheck \\
RGCL & & \faCheck & \faCheck & \faCheck & \faCheck & & \\
Rouges & \faCheck & \faCheck & \faCheck & \faCheck & \faCheck & & \\
RTNLU & & & & & \faCheck & & \\
SAFA & \faCheck & & & & & & \\
SaiSakethAluru & \faCheck & & & & & & \\
SAJA & \faCheck & & & & & & \\
saradhix & \faCheck & & & & & & \\
saroarj & \faCheck & & & & & & \\
SINAI & & & & & \faCheck & & \\
Smatgrisene & & \faCheck & & & & & \\
Sonal & & \faCheck & \faCheck & \faCheck & & & \\
sonal.kumari & \faCheck & & & & \faCheck & & \faCheck \\
SpurthiAH & \faCheck & & & & \faCheck & & \\
SRIB2020 & & \faCheck & & \faCheck & \faCheck & \faCheck & \faCheck \\
SSN\_NLP\_MLRG & & \faCheck & \faCheck & & \faCheck & \faCheck & \faCheck \\
Stormbreaker & & \faCheck & \faCheck & \faCheck & & & \\
SU-NLP & & & \faCheck & \faCheck & & & \\
Taha & \faCheck & & & & & & \\
TAC & \faCheck & \faCheck & \faCheck & \faCheck & \faCheck & & \\
GruPaTo & & & & \faCheck & \faCheck & & \\
Team Oulu & & \faCheck & \faCheck & & \faCheck & \faCheck & \faCheck \\
TeamKGP & & \faCheck & \faCheck & \faCheck & \faCheck & & \\
TECHSSN & & & & & \faCheck & \faCheck & \\
tharindu & \faCheck & & & & & & \\
TheNorth & & & & & \faCheck & & \\
TOBB ETU & & & & \faCheck & & & \\
TysonYU & \faCheck & \faCheck & \faCheck & \faCheck & \faCheck & \faCheck & \\
UHH-LT & & & & & \faCheck & \faCheck & \faCheck \\
UJNLP & & & & & \faCheck & & \\
\bottomrule
\end{tabular}
\caption{Overview of team participation in the subtasks (part 2).}
\label{table:team:participation:part2}
\end{table}

\begin{table}[tbh]
\small
\begin{tabular}{lccccccc}
\toprule
\bf Team & \bf A-Arabic & \bf A-Danish & \bf A-Greek & \bf A-Turkish & \bf A-English & \bf B-English & \bf C-English \\
\midrule
ultraviolet & & & & & & & \faCheck \\
UNT Linguistics & & & & & \faCheck & \faCheck & \\
UoB & & & & & \faCheck & \faCheck & \\
UTFPR & & & & & \faCheck & & \\
VerifiedXiaoPAI & & & & & \faCheck & \faCheck & \\
wac81 & & & & & \faCheck & \faCheck & \faCheck \\
will\_go & \faCheck & \faCheck & \faCheck & \faCheck & & & \\
KUISAIL & & \faCheck & \faCheck & & & & \\
Wu427 & & & & & \faCheck & \faCheck & \\
XD & & & & & \faCheck & & \\
yasserotiefy & \faCheck & & & & & & \\
yemen2016 & \faCheck & & & & & & \\
YNU\_oxz & & & & & \faCheck & & \\
zahra.raj & \faCheck & & & & & & \\
zoher\_orabe & \faCheck & & & & & & \\
\bottomrule
\end{tabular}
\caption{Overview of team participation in the subtasks (part 3).}
\label{table:team:participation:part3}
\end{table}

\end{document}